\documentclass[conference]{IEEEtran}
\IEEEoverridecommandlockouts
\usepackage{cite}
\usepackage{booktabs}       % professional-quality tables
\usepackage{amsfonts}       % blackboard math symbols
\usepackage{nicefrac}       % compact symbols for 1/2, etc.
\usepackage{microtype}      % microtypography
\usepackage{xcolor}         % colors
\usepackage{amsmath}
\usepackage{float}
\usepackage{graphicx}
\usepackage{tabularx}
\usepackage{multirow}
\usepackage{adjustbox}
\usepackage{caption}

\def\BibTeX{{\rm B\kern-.05em{\sc i\kern-.025em b}\kern-.08em
    T\kern-.1667em\lower.7ex\hbox{E}\kern-.125emX}}
\begin{document}

\title{Detecting Adversarial Images through Response Profiles of Vision-Language Models}

\author{Arash Vashagh and Roozbeh Razavi-Far\\
Trustworthy and Secure AI (TSAI) Lab, Faculty of Computer Science, University of New Brunswick, Canada\\
Email: arash.vashagh@unb.ca, roozbeh.razavi-far@unb.ca}

\maketitle

\begin{abstract}
Adversarial perturbations can alter the predictions of frozen
vision-language models (VLMs) while leaving their confidence and
image--text similarity patterns seemingly plausible. We investigate
whether we can identify adversarial inputs based on the broader
way an image interacts with a collection of general semantic prompts.
Our detector summarizes these responses using category-level statistics,
relationships among prompts, deviations from clean reference
distributions, and stability under weak image transformations, producing
a compact response profile that is classified by a lightweight model
while the VLM remains fixed. We evaluate the approach on multiple
public image datasets, several CLIP-style visual backbones, and a range
of gradient-based, optimization-based, automated, and spatial attacks.
The detector achieves strong discrimination in attack-specific settings
and retains substantial performance when evaluated on attacks not seen
during training. Under a controlled detector-specific protocol, the
response-profile representation outperforms the evaluated
embedding-geometry baselines. Additional analyses show that the feature
groups provide complementary information and that the method remains
effective under variations in the prompt configuration. We also examine
inference cost and performance against detector-aware adaptive attacks.
Overall, the results indicate that response patterns across semantic
prompts provide a useful complementary signal for adversarial image
detection in frozen VLMs.
\end{abstract}

\begin{IEEEkeywords}
adversarial examples, adversarial detection, vision-language models, CLIP, trustworthy machine learning
\end{IEEEkeywords}

\section{Introduction}

Vision-language models (VLMs) have become a central component of modern AI systems \cite{radford2021learning,li2021align,li2022blip}. Models such as CLIP \cite{radford2021learning} align images and natural-language descriptions in a shared representation space, enabling zero-shot recognition, multimodal retrieval, and open-vocabulary reasoning without task-specific retraining. These capabilities make VLM representations attractive for downstream systems that must interpret visual inputs through flexible semantic concepts. Their growing use in applications such as content moderation and autonomous systems \cite{zhou2023vision} also makes the reliability of these representations increasingly important.

Despite their strong generalization capabilities, VLMs remain vulnerable to adversarial perturbations. As with conventional vision models, small input perturbations can change the model prediction while leaving the image visually similar to the original \cite{szegedy2013intriguing,goodfellow2014explaining,madry2017towards,zhao2023evaluating}. Detecting such inputs is challenging because an adversarial example does not necessarily produce an obviously abnormal final confidence score. A successful perturbation may still yield a confident zero-shot prediction or apparently plausible image--text similarities. Consequently, examining only the final classifier output can discard information contained in the broader representation produced by the VLM.

Existing adversarial detectors exploit several forms of representation abnormality. Confidence-based approaches use quantities such as maximum softmax probability or energy scores \cite{hendrycks2016baseline,liu2020energy}. Other methods measure sensitivity to input transformations, density in feature space, local intrinsic dimensionality, Mahalanobis distance, or related representation statistics \cite{xu2017feature,ma2018characterizing,lee2018simple}. More recent VLM-specific methods analyze geometric properties of pretrained multimodal embeddings, attention responses, or learned semantic priors \cite{hasanebrahimi2025gadvlp,hasanebrahimi2026geodetect,zhang2024pip,10.1007/978-981-95-4367-0_17}. These approaches demonstrate that adversarial perturbations can leave detectable signatures beyond the final prediction.

Our motivation is that a frozen VLM provides another source of information that is naturally available but not fully represented by a single confidence score or embedding-space distance: its pattern of responses across many text prompts. Given an image and a collection of text prompts, a CLIP-style model produces a vector of image--text similarities rather than a single scalar output. This vector describes how the image is positioned relative to multiple directions in the joint vision--language space. Adversarial perturbations that alter the image representation can therefore change not only the predicted dataset class, but also the distribution, local relationships, reference deviations, and transformation behavior of these prompt-conditioned responses.

This observation motivates treating adversarial detection as a \emph{response-profiling} problem. Instead of asking only whether the input has low confidence or lies far from clean samples in the image embedding space, we characterize how the input interacts with a fixed semantic reference space. We use general prompts that are separate from the dataset class prompts used by the zero-shot classifier. Their image--text similarities provide a reusable representation from which we can extract several complementary measurements.

The novelty of our approach lies in constructing a compact detector representation from multiple properties of this prompt-conditioned response. First, category-level statistics summarize how similarity mass is distributed across broad semantic groups. Second, we use a graph built from text-embedding relationships to measure global and local variation among prompt responses. Third, clean-reference residuals quantify how the observed response differs from response distributions estimated from clean training images. Fourth, graph diffusion measures how the response changes under smoothing over related prompts. Finally, weak transformations of the input provide stability measurements that characterize how the VLM response changes under small image modifications. These components are concatenated into a compact response profile and classified by a lightweight multilayer perceptron.

This design differs from existing VLM-based adversarial detectors in both the representation analyzed and how the pretrained model is used. Geometry-based methods such as GeoDetect operate primarily on distances or density-related properties of pretrained embeddings \cite{hasanebrahimi2026geodetect}, while prompt-based irrelevant probing (PIP) uses attention responses generated by irrelevant probing questions \cite{zhang2024pip}. Our method instead operates directly on the pattern of image--text similarities generated by a fixed prompt bank and combines several summaries of that response into a single detector representation. The underlying image encoder, text encoder, prompt embeddings, and prompt graph remain frozen; only the small detector operating on the extracted features is trained.

An additional motivation for this design is interpretability at the feature-group level. Because the final detector operates on an explicitly constructed low-dimensional representation rather than a newly learned VLM representation, we can evaluate the contributions of different signal families separately. We can therefore remove or combine category, graph, residual, and transformation-stability components without modifying the frozen backbone. This lets us study the detector as a composition of measurable response properties rather than an opaque classifier operating directly on high-dimensional pretrained embeddings.

We evaluate the proposed detector across CIFAR-10, STL-10, and Tiny-ImageNet using ViT-B/16, ViT-B/32, and ConvNeXt CLIP-style backbones. The evaluation covers PGD, MI-FGSM, CW, AutoAttack, and flow-based attacks under both standard attack-specific and cross-attack settings. We additionally evaluate a confidence-based MSP detector, compare against embedding-geometry detectors under a controlled protocol, measure the contribution of individual feature groups through controlled ablations, quantify inference cost, and study detector-aware attacks. Together, these experiments evaluate both the effectiveness and generalization of the proposed response representation and the contribution of its individual components.

Our contributions are summarized as follows:
\begin{itemize}
    \item We introduce adversarial image detection through \emph{prompt-conditioned response profiling}, using the pattern of image--text similarities produced by a frozen VLM as the basis for detection.
    
    \item We develop a compact representation that combines category-level statistics, prompt-graph measurements, clean-reference residuals, graph-diffusion features, and transformation-stability features while keeping the underlying VLM fixed.
    
    \item We use a lightweight MLP detector that operates only on the extracted response profile, enabling analysis of individual feature groups without fine-tuning the VLM backbone.
    
    \item We provide a multi-dataset, multi-backbone, and multi-attack evaluation including standard and cross-attack detection, a confidence baseline, controlled comparisons with embedding-geometry detectors, feature ablations, runtime measurements, and detector-aware evaluation.
\end{itemize}

\section{Background}
This section reviews three groups of related work: adversarial example detection, VLM methods for semantic detection, and adversarial detection methods designed for VLMs. This context positions our prompt-conditioned response-profiling approach relative to existing confidence-, geometry-, and VLM-based detectors.

\subsection{Adversarial example detection}

Adversarial example detection has been widely studied for image classifiers. Early methods use confidence-based scores, including maximum softmax probability (MSP) and energy-based scoring, to identify abnormal inputs \cite{hendrycks2016baseline,liu2020energy}. Other approaches compare model behavior under input transformations, such as feature squeezing, or estimate sample density in deep feature spaces \cite{xu2017feature,feinman2017detecting}. Representation-based methods detect adversarial inputs using feature statistics or geometry, including local intrinsic dimensionality (LID), Mahalanobis distance, and statistical tests \cite{ma2018characterizing,lee2018simple,grosse2017statistical}. Nearest-neighbor methods further show that local representation structure can provide useful detection signals \cite{cohen2020detecting}. These methods can often identify adversarial inputs by detecting abnormal model behavior, but they are primarily designed for unimodal classifiers.

\subsection{VLMs for out-of-distribution and semantic detection}

VLMs provide semantic image-text representations that are useful for out-of-distribution (OOD) detection. CLIP aligns images and text in a shared embedding space \cite{radford2021learning}. Building on this, maximum concept matching (MCM) uses image-text similarity scores for OOD detection \cite{ming2022delving}. CLIPN introduces negation-aware prompts for zero-shot OOD rejection \cite{wang2023clipn}. Negative-label methods use auxiliary labels that describe concepts absent from the input or outside the target label set to improve semantic separation \cite{jiang2024negative,li2024learning,he2026out}, while SimLabel uses consistency across related labels \cite{zou2025simlabel}. Recent work further analyzes the mechanisms and sensitivities of VLM-based OOD detection \cite{lee2025empirical}, and surveys summarize progress with CLIP-like models \cite{li2025recent}.

These works show the value of semantic alignment for detecting distribution shifts. However, they mainly address OOD detection (not adversarial detection) and usually rely on class-specific prompts or aggregated image-text similarities.

\subsection{Adversarial detection with VLMs}

Recent work has studied adversarial detection directly in vision-language settings. GAD-VLP applies geometric measures to representations extracted from unimodal or multimodal vision-language pre-trained model (VLP) encoders \cite{hasanebrahimi2025gadvlp}. These measures include LID, k-NN distance, Mahalanobis distance, and kernel density estimation (KDE). GeoDetect studies similar geometric signals and connects adversarial detection to off-manifold behavior in anisotropic VLP embedding spaces \cite{hasanebrahimi2026geodetect}. PIP detects adversarial examples in large vision-language models (LVLMs) using attention patterns induced by irrelevant probe questions \cite{zhang2024pip}. Another recent method uses CLIP text embeddings as semantic priors to guide learned latent representations for adversarial detection \cite{10.1007/978-981-95-4367-0_17}.

These methods show that VLMs provide useful signals for adversarial detection. They mainly focus on embedding geometry, attention maps, or learned latent spaces. In contrast, our method analyzes image-text similarity patterns from semantic prompts that are not tied to dataset labels. It models prompt relationships with a graph and extracts category, residual, and stability features from a frozen VLM.

\subsection{Relationship to prior VLM-based detectors}

GAD-VLP and GeoDetect detect adversarial inputs from geometric deviations in VLP embedding spaces, using signals such as k-nearest-neighbor distance, Mahalanobis distance, KDE, and LID \cite{hasanebrahimi2025gadvlp,hasanebrahimi2026geodetect}. PIP instead probes large VLMs with irrelevant questions and classifies attention-derived responses \cite{zhang2024pip}. Other approaches use CLIP text embeddings as semantic priors for learned adversarial detectors \cite{10.1007/978-981-95-4367-0_17}.

Our method differs in the representation supplied to the detector. It operates on prompt-conditioned image--text responses and combines category-level statistics, prompt-graph measurements, clean-reference residuals, graph diffusion, and transformation stability. The VLM itself remains frozen, and only the lightweight detector is trained.

\section{Method}

This section outlines the proposed framework. First, we formally define the problem setup and specify the threat model. Subsequently, we establish the semantic anchor space, followed by the graph-structured semantic representation, semantic residual modeling, and semantic stability under small transformations. Finally, we provide a technical description of the final detector.

\subsection{Problem setup and notation}

Let $x \in \mathcal{X}$ denote an input image. The detector receives $x$ and outputs a score indicating whether the image is clean or adversarial. We use a pretrained vision-language model as a frozen feature extractor, with image encoder $f: \mathcal{X} \rightarrow \mathbb{R}^d$ and text encoder $g: \mathcal{P} \rightarrow \mathbb{R}^d$, where $\mathcal{P} = \{t_1, \dots, t_M\}$ is a set of semantic prompts, $M$ is the number of prompts, and $d$ is the embedding dimension.

The normalized image embedding is $z_I(x) = f(x)/\|f(x)\|_2$, and the normalized prompt embedding is $z_T(t_i) = g(t_i)/\|g(t_i)\|_2$. Stacking all prompt embeddings gives
\begin{equation}
\label{eq:text-embedding-matrix}
Z_T =
\begin{bmatrix}
z_T(t_1)^\top \\
\vdots \\
z_T(t_M)^\top
\end{bmatrix}
\in \mathbb{R}^{M \times d}.
\end{equation}

The semantic similarity vector is $s(x) = Z_T z_I(x)$, where $s(x) \in \mathbb{R}^M$.

\begin{figure*}[t]
\centering
\includegraphics[width=0.95\linewidth]{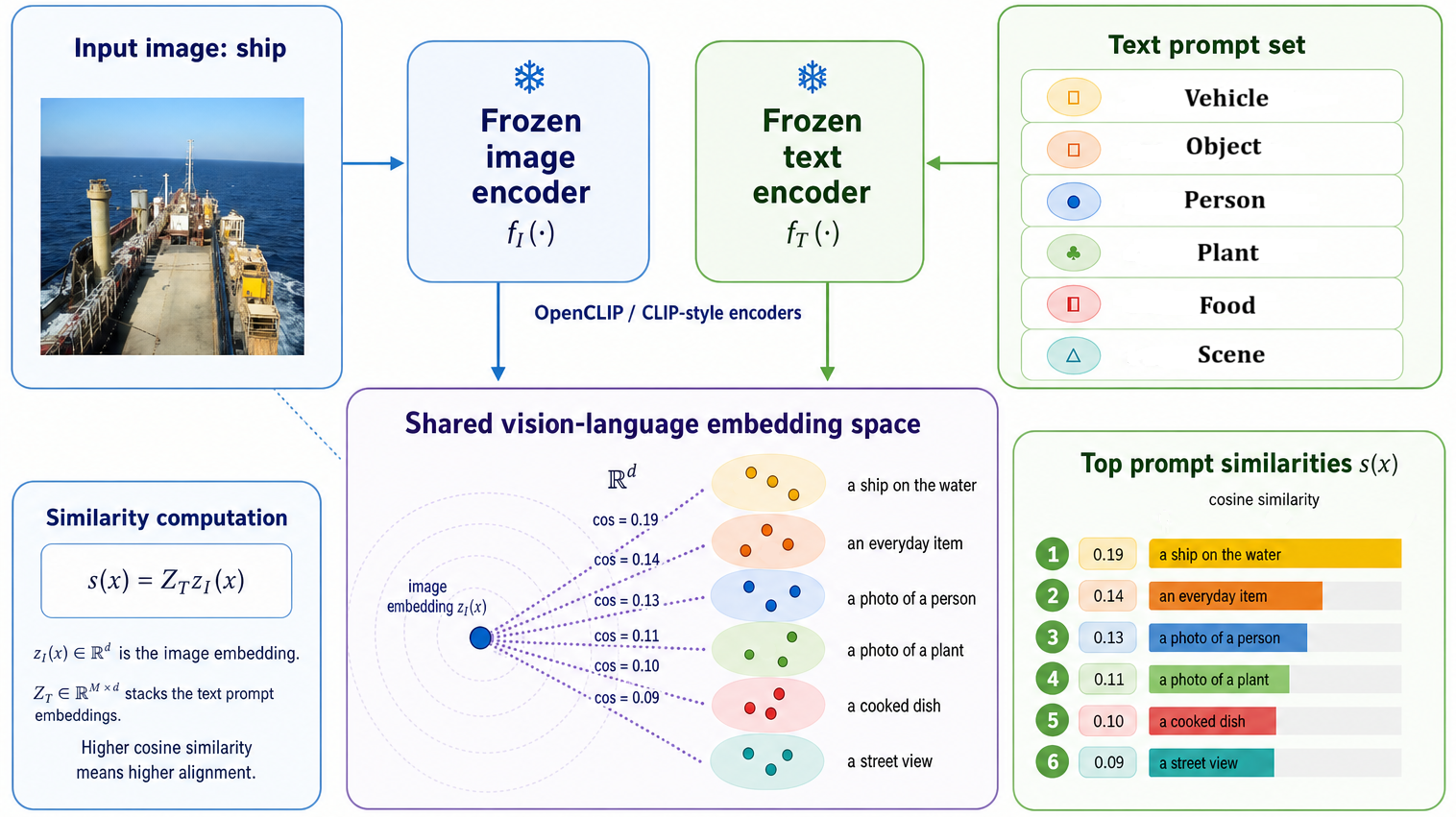}
\caption{Conceptual example of image--text similarity computation for a CIFAR-10 test sample labeled as a ship using a CLIP-style VLM with ViT-B/16 as the backbone. The image encoder maps the input to the normalized image embedding $z_I(x)$, while the text encoder maps the semantic prompts to text embeddings stacked in $Z_T$. The semantic response vector is computed as $s(x)=Z_T z_I(x)$ and contains one similarity value per prompt. For visualization, the figure shows only selected high-scoring prompts and their associated semantic categories rather than the complete prompt set. The full detector uses all seven semantic categories defined in the method.}
\label{fig:vlm_similarity}
\end{figure*}

Figure~\ref{fig:vlm_similarity} illustrates the frozen VLM similarity computation for one CIFAR-10 test image.

Prompts are grouped into $K$ semantic categories $\mathcal{C} = \{C_1, \dots, C_K\}$. From $s(x)$, we construct a feature vector $\phi(x) \in \mathbb{R}^{d_\phi}$, where $d_\phi$ denotes the feature dimension. We then train a lightweight multilayer perceptron (MLP) detector $D_\theta: \mathbb{R}^{d_\phi} \rightarrow \mathbb{R}$, where $\theta$ denotes the learnable MLP parameters. Higher detector scores indicate a higher likelihood of adversarial input.

\subsection{Threat model}

For each experiment, the target VLM uses a fixed visual backbone, such as ViT-B/16 \cite{dosovitskiy2020image}, ViT-B/32 \cite{dosovitskiy2020image}, or ConvNeXt \cite{liu2022convnet}. The selected backbone defines the CLIP-style VLM used in each experiment. We use this VLM to generate adversarial images by attacking its zero-shot image-text classification logits, and we also use the same frozen VLM to extract the semantic features used by the detector.

Let $\{c_1,\dots,c_N\}$ denote the dataset class names. To generate adversarial examples, the attacker constructs class-level prompts of the form ``a photo of a \{class\}.'' Let
\begin{equation}
\label{eq:class-prompt-matrix}
U =
\begin{bmatrix}
u_1^\top \\
\vdots \\
u_N^\top
\end{bmatrix}
\in \mathbb{R}^{N \times d}
\end{equation}
be the normalized text embeddings of these class-level prompts, where $N$ is the number of dataset classes and $u_i \in \mathbb{R}^d$ is the text embedding of the $i$-th class prompt.

The attack uses image-text logits $\ell(x) = \rho \, U z_I(x)$, where $\rho$ is the learned logit scale of the VLM if available.

Given a clean image $x$ with class label $y_{\mathrm{cls}}$, the adversary creates $x_{\mathrm{adv}}$ by optimizing a loss over these logits. For untargeted gradient-based attacks, this objective can be written as
\begin{equation}
\label{eq:attack-objective}
x_{\mathrm{adv}}
=
\arg\max_{x' \in \mathcal{B}(x)}
\mathcal{L}_{\mathrm{CE}}(\ell(x'), y_{\mathrm{cls}}),
\end{equation}
where $\mathcal{L}_{\mathrm{CE}}$ is the cross-entropy loss and $\mathcal{B}(x)$ is the allowed perturbation set around $x$.

The attacker has white-box access to the target VLM used for attack generation, including its image encoder, class-level text embeddings, logits, labels, and gradients. This access enables attacks such as PGD \cite{madry2017towards}, MI-FGSM \cite{dong2018boosting}, CW \cite{carlini2017towards}, AutoAttack \cite{croce2020reliable}, and flow-based attacks \cite{xiao2018spatially} to change the zero-shot classification scores by comparing image embeddings with dataset-class prompt embeddings.

\subsection{Semantic anchor space}

We aggregate prompt-level similarities into category-level statistics to summarize how the input aligns with broad semantic groups.

We normalize the similarity vector as $\hat{s}(x) = s(x)/\|s(x)\|_2$. For each category $C_k$, define:
\begin{equation}
\label{eq:category-statistics}
\begin{aligned}
m_k(x) &= \sum_{i \in C_k} \max(0, \hat{s}_i(x)),
&\;
\mu_k(x) &= \frac{1}{|C_k|} \sum_{i \in C_k} \hat{s}_i(x),\\
\sigma_k(x) &= \sqrt{\frac{1}{|C_k|} \sum_{i \in C_k} (\hat{s}_i(x) - \mu_k(x))^2},
&\;
\psi_k(x) &= \max_{i \in C_k} \hat{s}_i(x).
\end{aligned}
\end{equation}

where $m_k(x)$ measures the positive semantic mass of category $C_k$, while $\mu_k(x)$, $\sigma_k(x)$, and $\psi_k(x)$ summarize the average, spread, and strongest prompt response within that category.

To obtain a normalized semantic distribution, we define:
\begin{equation}
\label{eq:normalized-category-mass}
\tilde{m}_k(x) = \frac{m_k(x)}{\sum_{j=1}^K m_j(x) + \epsilon},
\end{equation}
where $\epsilon > 0$ is a small constant for numerical stability.

To measure how concentrated the semantic mass is across categories, we compute:
\begin{equation}
\label{eq:semantic-entropy}
H(x) = - \sum_{k=1}^K \tilde{m}_k(x) \log(\tilde{m}_k(x) + \epsilon).
\end{equation}

To capture separation between dominant categories, we define $\gamma(x) = \tilde{m}_{(1)}(x) - \tilde{m}_{(2)}(x)$, where $\tilde{m}_{(1)}(x)$ and $\tilde{m}_{(2)}(x)$ are the largest and second-largest normalized category masses. The entropy and margin summarize whether the semantic response is scattered (diffuse) across categories or concentrated around a dominant concept.

\subsection{Graph-structured semantic representation}

This step models relationships between prompts to capture structured semantic behavior. The similarity vector is treated as a signal over a graph defined on the prompt space, where each graph node corresponds to one semantic prompt.

We define a prompt similarity matrix $Q = Z_T Z_T^\top$. A sparse adjacency matrix $A \in \mathbb{R}^{M \times M}$ is constructed from $Q$ using top-$r$ nearest neighbors and then symmetrized, where $r$ is the number of graph neighbors. Let $G$ denote the diagonal degree matrix with entries $G_{ii}=\sum_{j=1}^M A_{ij}$. The graph Laplacian is $L=G-A$. To measure global smoothness of the semantic signal, we compute $E_G(x) = s(x)^\top L s(x)$. A low value of $E_G(x)$ indicates that related prompts have similar responses, while a high value indicates irregular semantic variation across the prompt graph. To capture local inconsistencies, we define node-wise deviations:
\begin{equation}
\label{eq:local-graph-deviation}
\delta_i(x) = s_i(x)^2 G_{ii} - 2 s_i(x) \sum_{j=1}^M A_{ij} s_j(x) + \sum_{j=1}^M A_{ij} s_j(x)^2.
\end{equation}

We summarize $\delta(x)$ using its maximum, mean, variance, and the mean of its $k_{\mathrm{graph}}$ largest values:
\begin{equation}
\label{eq:topkmean-delta}
\mathrm{topkmean}(\delta(x))
=
\frac{1}{k_{\mathrm{graph}}}
\sum_{i \in \mathcal{I}_{k_{\mathrm{graph}}}(x)}
\delta_i(x),
\end{equation}
where $\mathcal{I}_{k_{\mathrm{graph}}}(x)$ indexes the $k_{\mathrm{graph}}$ largest entries of $\delta(x)$. These statistics capture whether semantic irregularity is localized to a few prompts or spread across the graph. A visualization of the resulting prompt graph is provided in Section~\ref{sec:prompt_graph}.

\subsection{Semantic residual modeling}

This step measures how much the semantic response deviates from clean data distributions. It allows detection based on distributional inconsistency rather than similarity magnitude alone.

For each category $C_k$, we estimate a clean mean vector $\bar{s}_k$ and covariance matrix $\Sigma_k$ from clean training samples. The residual for category $C_k$ is defined as:
\begin{equation}
\label{eq:semantic-residual}
R_k(x) = (s_k(x) - \bar{s}_k)^\top \Sigma_k^{-1} (s_k(x) - \bar{s}_k),
\end{equation}
where $s_k(x)$ is the subvector of $s(x)$ restricted to prompts in $C_k$.

Let $k_1$ and $k_2$ be the indices of the two largest category masses. We define $R_{\mathrm{top1}}(x) = R_{k_1}(x)$ and $R_{\mathrm{top2}}(x) = R_{k_2}(x)$. These residuals measure how far the strongest semantic categories of the input deviate from clean-image statistics. We use the top two categories because adversarial perturbations may affect not only the dominant semantic category but also the nearest competing category.

\subsection{Semantic stability under small transformations}

This step evaluates how semantic representations change under small input transformations. The goal is to detect instability that does not affect visual appearance but alters semantic behavior. Let $\mathcal{A}$ be a set of small image transformations. For each transformation $a \in \mathcal{A}$, we compute the transformed image embedding $z_I(a(x))$ and its semantic response $s(a(x))$. To measure embedding stability, we compute the average cosine similarity between the original and transformed image embeddings:
\begin{equation}
\label{eq:embedding-stability}
S(x)
=
\frac{1}{|\mathcal{A}|}
\sum_{a \in \mathcal{A}}
z_I(x)^\top z_I(a(x)).
\end{equation}

To measure semantic drift under transformations, we compute:
\begin{equation}
\label{eq:semantic-drift}
\Delta_{\mathcal{A}}(x)
=
\frac{1}{|\mathcal{A}|}
\sum_{a \in \mathcal{A}}
\|s(x)-s(a(x))\|_2.
\end{equation}

We also propagate the semantic signal over the prompt graph using diffusion:
\begin{equation}
\label{eq:graph-diffusion}
s^{(q+1)}(x)
=
(1-\alpha)s^{(q)}(x)
+
\alpha G^{-1}A s^{(q)}(x),
\end{equation}
where $\alpha \in (0,1)$ controls the diffusion strength, $q$ indexes the diffusion iteration, and $s^{(0)}(x)=s(x)$.

After $T_{\mathrm{diff}}$ diffusion steps, we compute the diffusion deviation:
\begin{equation}
\label{eq:diffusion-deviation}
\Delta_{\mathrm{diff}}(x)
=
\|s^{(T_{\mathrm{diff}})}(x)-s(x)\|_2.
\end{equation}

We also compute the cosine similarity between the original and diffused semantic signals:
\begin{equation}
\label{eq:diffusion-cosine}
C_{\mathrm{diff}}(x)
=
\frac{
s(x)^\top s^{(T_{\mathrm{diff}})}(x)
}{
\|s(x)\|_2\|s^{(T_{\mathrm{diff}})}(x)\|_2+\epsilon
}.
\end{equation}

Finally, we compute $H_{\mathrm{diff}}(x)$ as the entropy of the category-level semantic masses after graph diffusion. The transformation-stability features measure sensitivity to small image changes, while the diffusion features measure how the semantic response behaves under graph-based smoothing.

\subsection{Feature representation and detector}

We combine all components into a unified feature vector that captures semantic alignment, graph structure, clean-distribution residuals, and stability under small transformations.

\begin{equation}
\label{eq:category-feature-vectors}
\begin{aligned}
\mathbf{m}(x)&=[m_1(x),\dots,m_K(x)], \quad
\boldsymbol{\mu}(x)=[\mu_1(x),\dots,\mu_K(x)],\\
\boldsymbol{\sigma}(x)&=[\sigma_1(x),\dots,\sigma_K(x)], \quad
\boldsymbol{\psi}(x)=[\psi_1(x),\dots,\psi_K(x)].
\end{aligned}
\end{equation}

The final feature vector is constructed by concatenating category-level, graph-level, diffusion, residual, stability, and normalized embedding-norm features:
\begin{equation}
\label{eq:final-feature-vector}
\begin{aligned}
\phi(x)=\big[
&\mathbf{m}(x),\boldsymbol{\mu}(x),\boldsymbol{\sigma}(x),
\boldsymbol{\psi}(x),H(x),\gamma(x),\\
&E_G(x),\max(\delta(x)),\mathrm{mean}(\delta(x)),
\mathrm{var}(\delta(x)),\\
&\mathrm{topkmean}(\delta(x)),
\Delta_{\mathrm{diff}}(x),C_{\mathrm{diff}}(x),
H_{\mathrm{diff}}(x),\\
&R_{\mathrm{top1}}(x),R_{\mathrm{top2}}(x),
S(x),\Delta_{\mathcal{A}}(x),\|z_I(x)\|_2
\big].
\end{aligned}
\end{equation}

\begin{figure*}[t]
\centering
\includegraphics[width=0.95\linewidth]{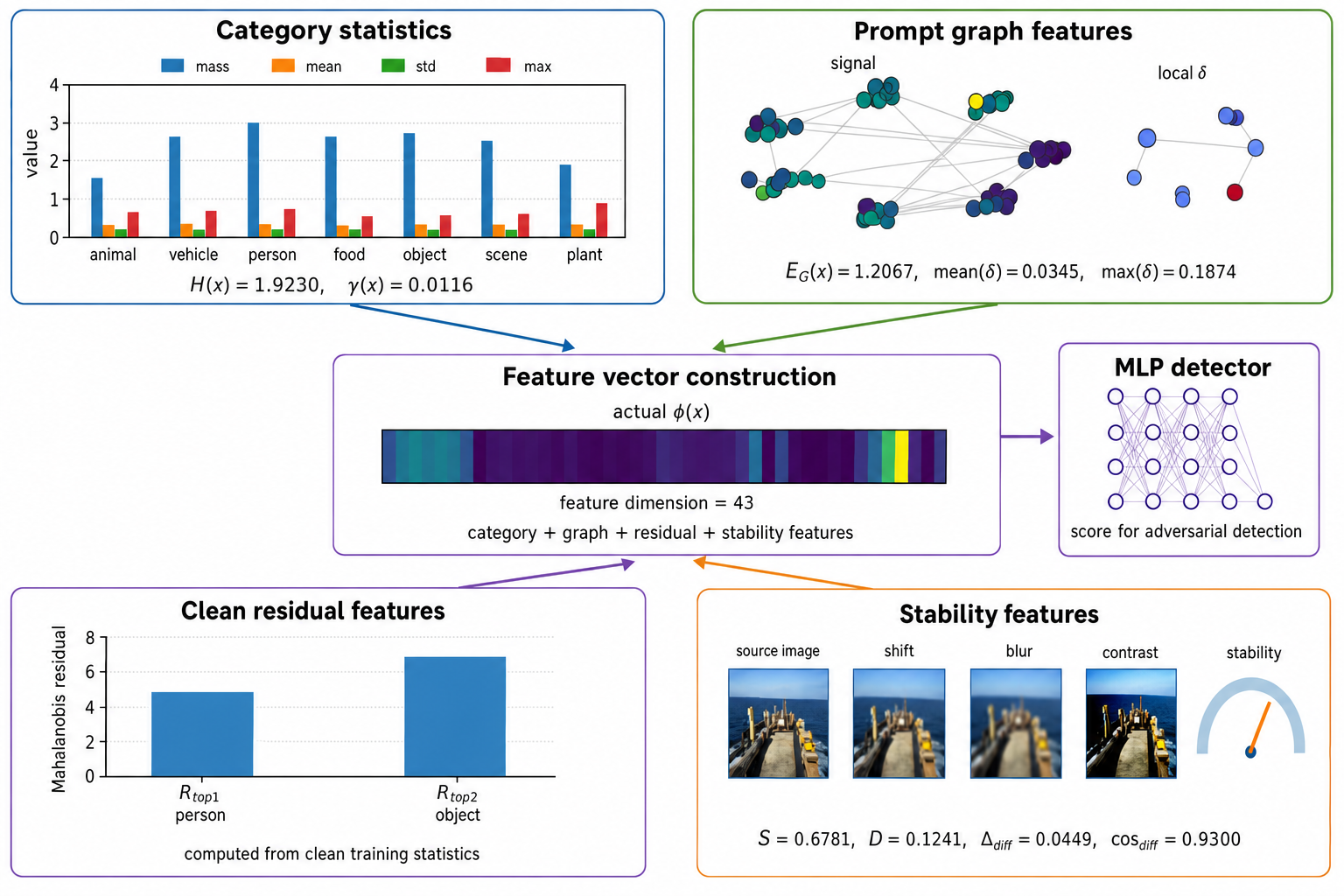}
\caption{
Example of feature-vector construction for a CIFAR-10 test image using a CLIP-style VLM with ViT-B/16 as backbone. The figure shows the actual feature groups computed by the pipeline for one sample: category statistics over the seven semantic prompt groups, prompt-graph features including graph energy and local deviations, clean-distribution residuals for the two dominant semantic categories, and stability features under small image transformations. These components are concatenated into the final feature vector $\phi(x)$ and passed to the MLP detector. For this example, the computed values include $H(x)=1.9230$, $\gamma(x)=0.0116$, $E_G(x)=1.2067$, $R_{\mathrm{top1}}=4.8525$, $R_{\mathrm{top2}}=6.8666$, $S(x)=0.6781$, and $\Delta_{\mathcal{A}}(x)=0.1241$.
}
\label{fig:feature_vector}
\end{figure*}

These components capture complementary effects of adversarial perturbations: category statistics summarize semantic alignment, graph features measure consistency among related prompts, residuals compare responses with clean-image statistics, and stability features measure sensitivity to small transformations. Together, they form a compact semantic profile. Figure~\ref{fig:feature_vector} illustrates the proposed feature-vector construction using the values computed for one CIFAR-10 test image. A breakdown of the $43$ feature dimensions is provided in Section~\ref{sec:feature_dimensionality}.

Before training the lightweight MLP detector $D_\theta$, we normalize the extracted features using statistics estimated from clean training samples. Let $\mu_\phi \in \mathbb{R}^{d_\phi}$ and $v_\phi \in \mathbb{R}^{d_\phi}$ denote the feature-wise mean and variance of $\phi(x)$ over clean training images. We define the whitened feature vector as
\begin{equation}
\label{eq:feature-whitening}
\phi_w(x) = \frac{\phi(x)-\mu_\phi}{\sqrt{v_\phi+\epsilon_w}},
\end{equation}
where $\epsilon_w>0$ is a small constant for numerical stability. The MLP detector is trained on $\phi_w(x)$ rather than the raw feature vector $\phi(x)$.

We train $D_\theta$ using $\phi_w(x)$ and binary detection labels $y_{\mathrm{det}} \in \{0,1\}$, where $y_{\mathrm{det}}=0$ denotes a clean image and $y_{\mathrm{det}}=1$ denotes an adversarial image. The VLM backbone and prompt embeddings remain fixed during training. The binary cross-entropy loss is:
\begin{equation}
\label{eq:bce-loss}
\begin{aligned}
\mathcal{L}_{\mathrm{BCE}}
=
-
\mathbb{E}\Big[
& y_{\mathrm{det}}
\log \operatorname{sigmoid}(D_\theta(\phi_w(x))) \\
& +
(1-y_{\mathrm{det}})
\log\!\left(
1-\operatorname{sigmoid}(D_\theta(\phi_w(x)))
\right)
\Big].
\end{aligned}
\end{equation}

To encourage adversarial samples to receive higher MLP scores than clean samples by at least a margin, we add:
\begin{equation}
\label{eq:margin-loss}
\mathcal{L}_{\mathrm{margin}} =
\mathbb{E} \left[
\max(0, \tau - D_\theta(\phi_w(x_{\mathrm{adv}})) + D_\theta(\phi_w(x_{\mathrm{clean}})))
\right],
\end{equation}
where $\tau > 0$ is the desired score margin between adversarial and clean samples.

The final objective for training the MLP detector is $\mathcal{L} = \mathcal{L}_{\mathrm{BCE}} + \lambda \mathcal{L}_{\mathrm{margin}}$, where $\lambda > 0$ controls the contribution of the margin loss. At inference time, the trained MLP outputs a detector score, and an input is classified as adversarial if $D_\theta(\phi_w(x)) \geq \eta$, where $\eta$ is a threshold selected on validation data.

The feature extraction procedure follows Eqs.~\eqref{eq:category-statistics}--\eqref{eq:final-feature-vector}: the frozen VLM produces prompt similarities, the semantic profile is constructed and whitened using clean-training statistics, and the MLP is optimized with the objective above. Computational complexity is discussed in Section~\ref{sec:computation}, while implementation details, attack parameters, and hardware are given in Sections~\ref{sec:implementation_details} and~\ref{sec:hardware}.

\subsection{Computational complexity}
\label{sec:computation}
Let $r$ denote the number of graph neighbors, $T_{\mathrm{diff}}$ the number of diffusion steps, $|\mathcal{A}|$ the number of input transformations, and $N_{\mathrm{tr}}$ the number of training samples. The quantities $M$, $d$, and $d_\phi$ are defined above.

The inference complexity is
\begin{equation}
\mathcal{O}\big((1 + |\mathcal{A}|)\, C_{\mathrm{model}}
+ |\mathcal{A}| \cdot (M d + M r)
+ T_{\mathrm{diff}} \cdot M r \big).
\end{equation}
where $C_{\mathrm{model}}$ is the cost of a single forward pass of the vision-language model. The first term accounts for computing image embeddings for the original input and transformed views. The second term corresponds to similarity computation and graph-based features across transformations. The third term captures diffusion on the prompt graph. In practice, $C_{\mathrm{model}}$ dominates, yielding $\mathcal{O}((1 + |\mathcal{A}|)\, C_{\mathrm{model}})$.

The training complexity is
\begin{equation}
\mathcal{O}\big(N_{\mathrm{tr}} \cdot [(1 + |\mathcal{A}|)\, C_{\mathrm{model}} + d_\phi]\big).
\end{equation}

This reflects that the backbone remains fixed during training and the trainable component operates only on the extracted feature vector.

\section{Experimental setup}
\label{sec:experimental_setup}

We evaluate the proposed detector across multiple image datasets, frozen VLM backbones, and adversarial attack families under standard, cross-attack, controlled-baseline, ablation, and detector-aware settings. This section summarizes the implementation choices, feature construction, detector configuration, attack parameters, and computational environment used throughout the experiments.

\subsection{Implementation details}
\label{sec:implementation_details}

This section provides additional details about the prompt-graph construction, clean residual statistics, feature normalization, image transformations, detector architecture, attack settings, and computational complexity in our experiments.

\subsubsection{Prompt graph construction}
\label{sec:prompt_graph}
\begin{figure*}[t]
\centering
\includegraphics[width=0.95\linewidth]{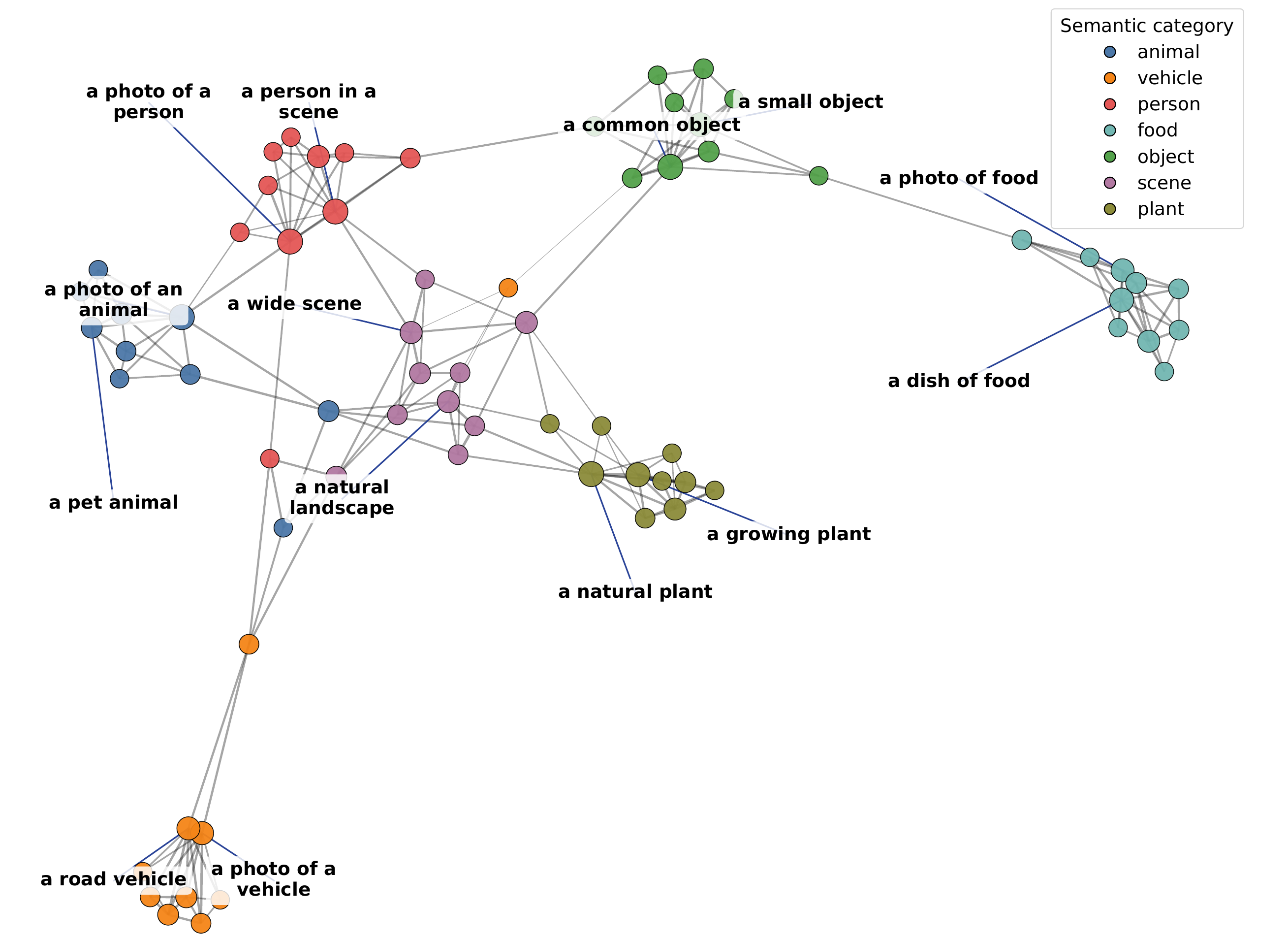}
\caption{
Visualization of the prompt graph used in the proposed detector. Nodes correspond to semantic prompts and are colored by semantic category. Black edges connect top-$r$ nearest neighbors in the frozen text-embedding space and are symmetrized before graph construction. Node size is proportional to graph degree. Blue guide lines connect selected prompt labels to their corresponding nodes and are included only for readability; they are not graph edges. The graph provides the fixed structure used to compute graph-based response features.
}
\label{fig:prompt_graph}
\end{figure*}

Prompt embeddings are computed once with the frozen text encoder and reused for all samples. We construct the adjacency matrix by connecting each prompt to its top-$r$ nearest neighbors in normalized text-embedding space. We then symmetrize the adjacency matrix before computing the graph Laplacian. This graph is fixed after construction and is shared across all images for a given prompt set. Figure~\ref{fig:prompt_graph} visualizes the resulting prompt graph.

\subsubsection{Clean residual statistics and feature whitening}

Clean residual statistics are estimated only from clean training samples. For each semantic category, we compute the clean mean vector and covariance matrix of the corresponding prompt-similarity subvector. These statistics are then used to compute residual features for validation and test samples. In practice, covariance shrinkage is used with shrinkage coefficient $0.20$ to improve numerical stability when estimating category-level covariance matrices.

Before training the detector, feature whitening is applied to the final feature vector. The feature-wise mean $\mu_\phi$ and variance $v_\phi$ are estimated from clean training samples, and each feature vector is normalized as
\begin{equation}
\phi_w(x) = \frac{\phi(x)-\mu_\phi}{\sqrt{v_\phi+\epsilon_w}},
\end{equation}
where $\epsilon_w$ is a small constant for numerical stability. This step prevents features with larger numerical ranges from dominating the MLP detector and improves training stability.

\subsubsection{Transformations and hyperparameters}

We use lightweight transformations to measure semantic stability. These include small spatial shifts, mild blur, and minor photometric changes such as contrast adjustment. The transformations are intentionally weak so that the main semantic content of the image is preserved. Stronger transformations may alter the meaning of clean images, while transformations that are too weak may fail to reveal instability introduced by adversarial perturbations.

The graph diffusion strength is set to $\alpha = 0.15$ to smooth the
semantic signal over the prompt graph while preserving local
variations. The margin-loss weight is set to $\lambda = 10^{-3}$.
These values are chosen to balance detection performance,
numerical stability, and computational efficiency.

\subsubsection{Feature-vector dimensionality}
\label{sec:feature_dimensionality}

The final feature vector has dimension $d_\phi=43$ in all experiments. This follows from the use of $K=7$ semantic categories and $15$ additional scalar features. Specifically, the category-level statistics consist of four vectors for each semantic category: positive semantic masses, category means, category standard deviations, and category maxima. These contribute $4K=28$ features.

The remaining $15$ scalar features are:
semantic entropy $H(x)$, semantic margin $\gamma(x)$, global graph energy $E_G(x)$, four localized graph-anomaly statistics, three graph-diffusion and graph-entropy statistics, two clean-distribution residual scores, two transformation-stability scores, and the image-embedding norm. Therefore,
\begin{equation}
d_\phi = 4K + 15 = 4 \times 7 + 15 = 43.
\end{equation}

In our implementation, the localized graph-anomaly statistics are the maximum, mean, variance, and top-$k$ mean of the node-wise graph deviations. The graph-diffusion and graph-entropy statistics are the diffusion deviation, diffusion cosine similarity, and graph entropy after diffusion. The two clean-distribution residual scores correspond to the residuals of the two categories with the largest semantic masses. The two transformation-stability scores measure semantic stability under small input transformations.

\subsubsection{Detector architecture}

\begin{figure*}[t]
\centering
\includegraphics[width=0.95\linewidth]{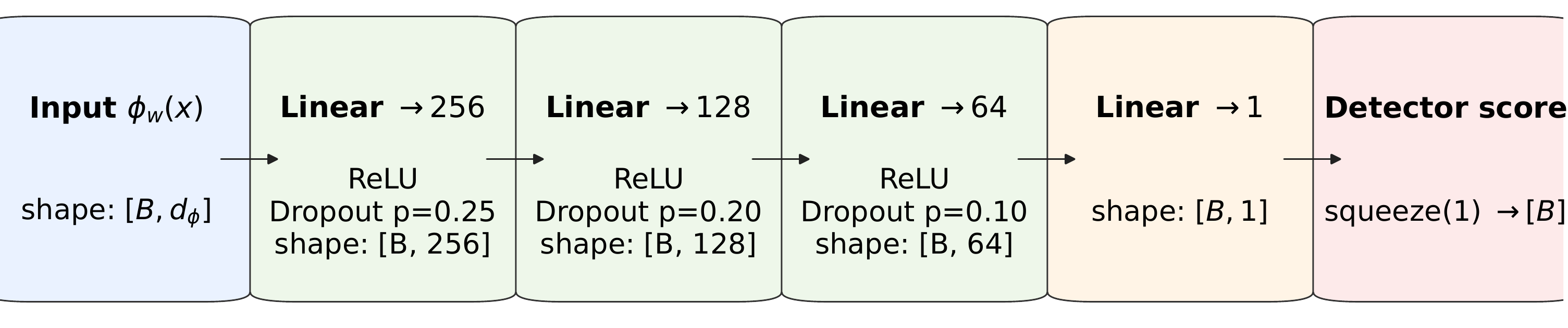}
\caption{
Architecture of the lightweight MLP detector. The input is the whitened response-profile feature vector $\phi_w(x) \in \mathbb{R}^{d_\phi}$. The detector uses three hidden layers with dimensions $256$, $128$, and $64$, each followed by ReLU activation and dropout, and a final linear layer that outputs a scalar detector score. Higher scores indicate a higher likelihood of adversarial input.
}
\label{fig:mlp_detector_arch}
\end{figure*}

The detector is a lightweight multilayer perceptron trained on the whitened feature vector $\phi_w(x)$. In our implementation, the MLP has hidden dimensions $256$, $128$, and $64$, with ReLU activations and dropout after each hidden layer. The final layer outputs a single scalar detector score. The compact architecture combines the low-dimensional response-profile features without fine-tuning the VLM backbone. The VLM backbone, prompt embeddings, prompt graph, and clean residual statistics remain fixed during detector training. Figure~\ref{fig:mlp_detector_arch} illustrates the detector architecture.

\subsubsection{Attack settings}

For the standard and cross-attack evaluations in Section~\ref{sec:standard_transfer}, we use five adversarial attacks: PGD, MI-FGSM, CW, AutoAttack, and Flow. All attacks optimize the zero-shot CLIP-style image-text classification objective constructed from dataset class prompts of the form ``a photo of a \{class\}.''

For PGD and MI-FGSM, we use untargeted $\ell_\infty$-bounded attacks with random initialization. For CIFAR-10 and Tiny-ImageNet, we use perturbation budget $\epsilon_{\mathrm{atk}}=0.04$, step size $\alpha_{\mathrm{atk}}=0.01$, and $20$ iterations. For STL-10, we use $\epsilon_{\mathrm{atk}}=0.03$, $\alpha_{\mathrm{atk}}=0.0075$, and $10$ iterations. MI-FGSM additionally uses momentum factor $\mu_{\mathrm{mom}}=1.0$ with $\ell_1$-normalized gradients.

The CW attack is implemented with $200$ optimization steps, a learning rate of $0.005$, a confidence parameter of $\kappa=10$, and a trade-off coefficient of $c=5.0$. It optimizes a sum of squared perturbation magnitude and a margin-based classification loss over the CLIP-style image-text logits.

The Flow attack learns a dense spatial deformation field and warps the image using bilinear sampling. It uses the same dataset-dependent $\epsilon_{\mathrm{atk}}$, $\alpha_{\mathrm{atk}}$, and iteration count as PGD and MI-FGSM. The flow field is initialized randomly, clipped to the allowed range, and optimized using a smoothness term with a weight of $0.05$.

For AutoAttack, we use the same dataset-dependent $\ell_\infty$ perturbation budget as above, along with a batch size of $32$. On CIFAR-10 and STL-10, we use the standard AutoAttack evaluation protocol. On Tiny-ImageNet, we use the implementation's iterative fallback mode with $20$ steps and step size $2/255$ in pixel space.

\subsection{Hardware and runtime}
\label{sec:hardware}

The experiments were run using NVIDIA H100 NVL and NVIDIA A100 GPUs, depending on resource availability. Most large experimental batches were executed on a workstation equipped with two NVIDIA H100 NVL GPUs, each with approximately $95.8$GB of memory. Several additional runs, including diagnostic and ablation runs, were executed on an NVIDIA A100 GPU through a Colab-based environment.

The standard attack-specific evaluation consisted of $45$ model-dataset-attack configurations, corresponding to $3$ datasets, $3$ VLM backbones, and $5$ attacks. Additional experiments included cross-attack evaluation, controlled baseline comparisons, ablations, runtime measurements, and detector-aware evaluation. The main experimental runs took approximately 70 hours across the available GPU resources. Runtime varies across datasets, backbones, and attacks. Attacks with iterative optimization, such as CW, are more computationally expensive than shorter iterative attacks such as PGD and MI-FGSM. The dominant computational cost is the frozen VLM forward pass over the original and transformed views. Training the MLP detector is comparatively inexpensive because it operates only on the extracted feature vectors.

\section{Results}
\label{sec:results}

This section evaluates the proposed detector from several complementary
perspectives, including standard and cross-attack detection, controlled
comparisons with geometric baselines, inference-time overhead, feature
ablations, and detector-aware adaptive evaluation.

\subsection{Standard and cross-attack evaluation}
\label{sec:standard_transfer}

We evaluate the proposed detector on CIFAR-10, STL-10, and
Tiny-ImageNet using three CLIP-style VLM backbones: ViT-B/16,
ViT-B/32, and ConvNeXt. Five attack families are considered: PGD,
CW, MI-FGSM, Flow, and AutoAttack. For each dataset, we construct fixed,
disjoint training, validation, and test partitions using a
$70\%/15\%/15\%$ split, and the same partitions are used across
detector configurations.

In the standard setting, a separate detector is trained and evaluated
for each attack family using adversarial examples generated by that
attack on disjoint training, validation, and test samples. In the
cross-attack setting, the detector is trained using PGD adversarial
examples and then evaluated on CW, MI-FGSM, Flow, and AutoAttack
examples without retraining.

Table~\ref{tab:rocauc} reports ROC-AUC on combined clean and adversarial
test samples for both settings across all three datasets and backbones.

\begin{table}[t]
\centering
\caption{ROC-AUC (\%) on clean + adversarial samples.
Left: standard attack-specific evaluation.
Right: cross-attack evaluation with detectors trained on PGD.}
\label{tab:rocauc}
\setlength{\tabcolsep}{2.5pt}
\resizebox{\linewidth}{!}{%
\begin{tabular}{lcccccc}
\toprule
& \multicolumn{3}{c}{\textbf{Standard}}
& \multicolumn{3}{c}{\textbf{Cross-attack}} \\
\cmidrule(lr){2-4} \cmidrule(lr){5-7}
Attack
& ViT-B/16 & ViT-B/32 & ConvNeXt
& ViT-B/16 & ViT-B/32 & ConvNeXt \\
\midrule

\multicolumn{7}{c}{\textbf{CIFAR-10}} \\
PGD        & 98.87 & 97.78 & 98.96 & --    & --    & -- \\
CW         & 99.59 & 97.68 & 99.95 & 90.35 & 92.40 & 96.90 \\
MI-FGSM    & 98.58 & 97.93 & 98.53 & 95.07 & 95.00 & 92.39 \\
Flow       & 99.91 & 99.83 & 99.93 & 96.60 & 94.80 & 92.06 \\
AutoAttack & 99.01 & 98.56 & 98.86 & 97.84 & 96.20 & 97.81 \\

\midrule
\multicolumn{7}{c}{\textbf{STL-10}} \\
PGD        & 98.02 & 96.60 & 98.67 & --    & --    & -- \\
CW         & 98.23 & 97.53 & 99.68 & 93.40 & 92.10 & 97.09 \\
MI-FGSM    & 96.98 & 95.38 & 98.38 & 94.06 & 93.40 & 94.11 \\
Flow       & 99.48 & 99.22 & 99.76 & 92.30 & 91.70 & 90.13 \\
AutoAttack & 99.44 & 99.31 & 99.46 & 98.52 & 93.00 & 98.26 \\

\midrule
\multicolumn{7}{c}{\textbf{Tiny-ImageNet}} \\
PGD        & 98.15 & 94.65 & 98.97 & --    & --    & -- \\
CW         & 99.52 & 91.68 & 99.40 & 98.71 & 91.50 & 97.70 \\
MI-FGSM    & 99.77 & 94.65 & 99.95 & 99.50 & 92.80 & 99.00 \\
Flow       & 99.68 & 99.17 & 99.37 & 96.56 & 91.30 & 90.13 \\
AutoAttack & 99.87 & 91.04 & 99.83 & 99.81 & 92.20 & 97.61 \\

\bottomrule
\end{tabular}}
\end{table}

The detector achieves consistently high discrimination in the standard
setting across datasets, attacks, and VLM backbones. Cross-attack
performance also remains strong, indicating that a detector trained
using PGD examples can retain substantial discriminative ability on
attack families not observed during training.

\begin{figure}[t]
\centering
\includegraphics[width=\linewidth]{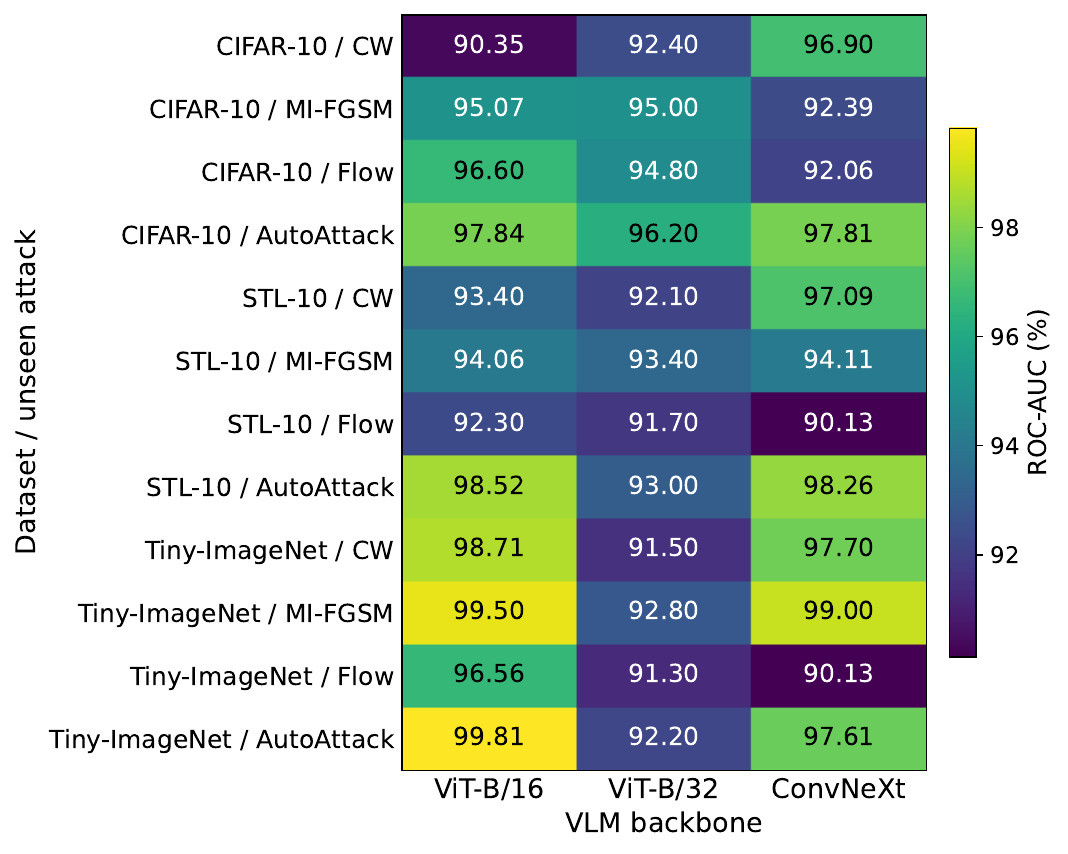}
\caption{Cross-attack ROC-AUC (\%) for detectors trained on PGD
adversarial examples and evaluated on unseen attack families.
Results are shown across three datasets and three VLM backbones.
Higher values indicate stronger transfer of the learned response-profile
detector across attacks.}
\label{fig:cross_attack_heatmap}
\end{figure}

Figure~\ref{fig:cross_attack_heatmap} provides a compact visualization
of the cross-attack results across datasets, unseen attacks, and VLM
backbones.

The cross-attack results are strongest for AutoAttack across most
configurations and remain consistently high for MI-FGSM. Transfer to CW
and Flow is also maintained across the evaluated backbones. Overall,
the results indicate that the detector does not depend exclusively on
attack-specific artifacts from the PGD training distribution.

\subsection{Controlled comparison with geometric baselines}
\label{sec:controlled_comparisons}

We compare the proposed detector with four representative
embedding-geometry detectors: k-nearest neighbors (KNN), Mahalanobis
distance, kernel density estimation (KDE), and local intrinsic
dimensionality (LID). The comparison uses CIFAR-10 and STL-10 with the
same frozen ViT-B/16-quickgelu encoder, data splits, sample budgets,
attack settings, and validation procedure for all methods.

For Table~\ref{tab:controlled_comparison}, each detector is evaluated
independently under PGD, CW, and Flow using the same detector-specific
evaluation protocol. Adversarial examples are generated separately for
each detector, and the attack procedure includes detector-aware
optimization. The final column reports the mean ROC-AUC across the three
attack families.

\begin{table}[t]
\centering
\caption{Controlled comparison with geometric detectors.
Entries are ROC-AUC (\%), and the final column reports mean ROC-AUC
across PGD, CW, and Flow.}
\label{tab:controlled_comparison}
\scriptsize
\setlength{\tabcolsep}{10pt}
\begin{adjustbox}{max width=\linewidth}
\begin{tabular}{llcccc}
\toprule
Dataset & Detector & PGD & CW & Flow & Mean \\
\midrule

\multirow{5}{*}{CIFAR-10}
& KNN           & 73.5 & 70.8 & 63.2 & 69.2 \\
& Mahalanobis   & 74.1 & 69.9 & 64.0 & 69.3 \\
& KDE           & 72.8 & 68.7 & 65.1 & 68.9 \\
& LID           & 68.4 & 66.8 & 61.7 & 65.6 \\
& Proposed      & \textbf{82.4} & \textbf{78.6} & \textbf{76.9}
                & \textbf{79.3} \\

\midrule

\multirow{5}{*}{STL-10}
& KNN           & 77.6 & 74.9 & 69.0 & 73.8 \\
& Mahalanobis   & 76.8 & 74.1 & 68.6 & 73.2 \\
& KDE           & 78.4 & 76.2 & 71.1 & 75.2 \\
& LID           & 67.0 & 64.8 & 62.7 & 64.8 \\
& Proposed      & \textbf{86.0} & \textbf{81.4} & \textbf{84.1}
                & \textbf{83.8} \\

\bottomrule
\end{tabular}
\end{adjustbox}
\end{table}

The proposed detector achieves the strongest mean ROC-AUC on both
datasets. These results
suggest that combining prompt-conditioned category statistics, graph
structure, clean-reference residuals, diffusion behavior, and
transformation stability provides complementary information to direct
embedding-space measurements.

\subsection{Inference-time overhead}
\label{sec:inference_overhead}

We measure batch-size-one, end-to-end GPU inference on an NVIDIA A100.
Measurements include the image encoder, feature construction, and
detector, while excluding data loading, model loading, and one-time
calibration. The full detector processes the original image together
with three weakly transformed views and therefore requires four
image-encoder passes. We additionally evaluate a one-view configuration
that processes only the original image and does not compute
transformation-stability features.

\begin{table}[t]
\centering
\captionsetup{justification=raggedright,singlelinecheck=false}
\caption{Batch-size-one inference latency and throughput on an NVIDIA A100.}
\label{tab:inference_overhead}
\setlength{\tabcolsep}{4pt}
\begin{tabular}{llcr}
\toprule
Dataset & Method & Latency (ms/image) & FPS \\
\midrule

\multirow{5}{*}{CIFAR-10}
& MSP                   & 5.88  & 170.10 \\
& GeoDetect-KNN         & 6.01  & 166.50 \\
& GeoDetect-Mahalanobis & 6.12  & 163.30 \\
& Proposed, one view    & 12.83 & 77.97 \\
& Proposed, four views  & 29.27 & 34.16 \\

\midrule

\multirow{5}{*}{STL-10}
& MSP                   & 5.84  & 171.14 \\
& GeoDetect-KNN         & 6.15  & 162.68 \\
& GeoDetect-Mahalanobis & 6.19  & 161.52 \\
& Proposed, one view    & 12.89 & 77.56 \\
& Proposed, four views  & 30.05 & 33.28 \\

\bottomrule
\end{tabular}
\end{table}

The full configuration requires additional computation because the frozen
image encoder is evaluated on four views of each input. The one-view
configuration reduces the number of encoder evaluations, while the
lightweight MLP itself contributes only a small portion of the total
inference cost.

\subsection{MSP-based baseline}
\label{sec:msp_baseline}

We compare against a confidence-based detector derived from maximum
softmax probability (MSP), a widely used baseline for identifying
abnormal inputs \cite{hendrycks2016baseline}. The baseline follows a
CLIP-style zero-shot classification setting in which class probabilities
are obtained from image--text similarity logits
\cite{radford2021learning}.

The baseline uses class-level prompts of the form
``a photo of a \{class\}'' and computes a softmax distribution over the
dataset classes. Given image--text logits $z_i$, the class probabilities
are

\begin{equation}
p_i =
\frac{\exp(z_i/T)}
{\sum_j \exp(z_j/T)},
\end{equation}

where $T$ is a temperature parameter. The detection score is $\mathrm{MSP}(x)=\max_i p_i$.

We select the detection threshold using validation data.
Table~\ref{tab:msp_auc} reports ROC-AUC across datasets, backbones, and
attacks.

\begin{table}[t]
\centering
\caption{ROC-AUC (\%) of MSP-based detection across datasets and models.}
\label{tab:msp_auc}
\setlength{\tabcolsep}{10pt}
\begin{tabular}{lccc}
\toprule
\textbf{Dataset / Attack}
& \textbf{ViT-B/16}
& \textbf{ViT-B/32}
& \textbf{ConvNeXt} \\
\midrule

\multicolumn{4}{c}{\textbf{CIFAR-10}} \\
PGD        & 89.95 & 91.68 & 94.04 \\
MI-FGSM    & 92.59 & 93.86 & 93.79 \\
CW         & 70.39 & 65.89 & 73.02 \\
Flow       & 94.04 & 93.69 & 95.83 \\
AutoAttack & 94.37 & 94.07 & 94.88 \\

\midrule
\multicolumn{4}{c}{\textbf{STL-10}} \\
PGD        & 67.43 & 69.23 & 78.18 \\
MI-FGSM    & 69.69 & 68.83 & 78.54 \\
CW         & 62.62 & 58.27 & 74.23 \\
Flow       & 74.82 & 72.27 & 79.27 \\
AutoAttack & 88.20 & 89.99 & 91.22 \\

\midrule
\multicolumn{4}{c}{\textbf{Tiny-ImageNet}} \\
PGD        & 99.42 & 89.42 & 89.87 \\
MI-FGSM    & 99.88 & 93.61 & 94.95 \\
CW         & 99.98 & 99.50 & 98.77 \\
Flow       & 99.45 & 90.31 & 68.05 \\
AutoAttack & 99.96 & 96.99 & 95.27 \\

\bottomrule
\end{tabular}
\end{table}

MSP performance varies substantially across datasets and attacks.
In contrast, the proposed detector maintains consistently high ROC-AUC
across the standard configurations in Table~\ref{tab:rocauc}, indicating
that the response-profile representation provides a more stable
detection signal than classification confidence alone.

\subsection{Ablation study}
\label{sec:ablation}

We conduct an ablation study on CIFAR-10 with the ViT-B/16 backbone
under PGD. Each variant uses the same training, validation, and test
protocol, and only the feature groups provided to the MLP detector are
changed. This experiment measures the contribution of the major
components of the response-profile representation.

Table~\ref{tab:ablation} shows that category-level features alone provide
a substantial detection signal. Adding graph features improves both
ROC-AUC and TPR at $5\%$ FPR, indicating that relationships among prompt
responses provide additional information beyond category-level
statistics. Residual features also improve over the category-only
representation.

Transformation-stability features provide the largest individual
improvement over category-level features. Combining category, graph,
and stability features further improves performance, while the complete
feature representation achieves the strongest overall result. The
ablation therefore indicates that the detector benefits from
complementary response properties rather than from a single feature
group.

\begin{table}[t]
\centering
\caption{Ablation study on CIFAR-10 with ViT-B/16 under PGD.
Dim.\ denotes feature dimension, Acc.\ denotes accuracy, and TPR@5
denotes the true-positive rate at $5\%$ false-positive rate.
Acc., F1, AUC, and TPR@5 are reported in percent.}
\label{tab:ablation}
\setlength{\tabcolsep}{5pt}
\begin{tabular}{lccccc}
\toprule
\textbf{Variant}
& \textbf{Dim.}
& \textbf{Acc.}
& \textbf{F1}
& \textbf{AUC}
& \textbf{TPR@5} \\
\midrule

Category only
& 30 & 87.08 & 87.30 & 94.69 & 74.10 \\

Category + graph
& 38 & 90.75 & 90.82 & 96.71 & 83.95 \\

Category + residual
& 32 & 88.12 & 88.05 & 95.29 & 76.45 \\

Category + stability
& 32 & 92.10 & 92.05 & 97.78 & 88.25 \\

Category + graph + residual
& 40 & 90.90 & 90.76 & 97.18 & 86.55 \\

Category + graph + stability
& 40 & 93.60 & 93.53 & 98.55 & 92.20 \\

Full model
& 43 & 94.08 & 93.94 & 98.70 & 92.85 \\

\bottomrule
\end{tabular}
\end{table}

\section{Discussion and future work}
\label{sec:discussion}

The detector uses $70$ general prompts organized into seven broad categories:
\emph{animal, vehicle, person, food, object, scene,} and \emph{plant}. The goal
is not to reproduce dataset labels, but to provide a reusable semantic
reference space from which multiple properties of the VLM response can be
measured. The ablation results show that category-level statistics already
provide a substantial detection signal, while graph, residual, and
transformation-stability features provide complementary information.
Transformation stability gives the largest individual improvement, and the
complete response-profile representation achieves the strongest overall
performance.

The cross-attack results further show that detectors trained on PGD retain
strong discrimination against CW, MI-FGSM, Flow, and AutoAttack across the
evaluated VLM backbones. In the controlled comparison, the proposed
response-profile representation also achieves higher mean ROC-AUC than the
evaluated embedding-geometry detectors under a common protocol. Together,
these results support the use of prompt-conditioned response profiling as a
detection signal that captures information beyond classification confidence
and direct embedding-space geometry.

The transformed views contribute strongly to detection performance while
also accounting for most of the additional inference cost measured in
Section~\ref{sec:inference_overhead}. The detector-aware evaluation further shows that the response-profile
representation retains substantial discriminative ability under adaptive
optimization.

Future work can explore automatic prompt construction and weighting, more
efficient transformation-stability features, broader multimodal
architectures, real-world distribution shifts, and additional forms of
detector-aware training and adaptive evaluation.

\section{Conclusion}

We presented a semantic-response profiling framework for adversarial image detection with frozen vision-language models. The detector summarizes image--text responses using category, graph, residual, diffusion, and transformation-stability features while leaving the VLM backbone fixed. Across multiple datasets, backbones, and attacks, the method achieves strong standard detection performance and strong cross-attack transfer. Controlled comparisons show that the proposed response-profile representation provides information beyond confidence-based and geometric embedding scores, while feature ablations demonstrate complementary contributions from the different feature groups, with transformation stability providing the largest individual improvement. Detector-aware evaluation further shows that the detector retains substantial discrimination under adaptive optimization. Overall, these results support semantic-response profiling as an effective complementary signal for adversarial image detection in frozen VLMs.

\bibliographystyle{IEEEtran}
\bibliography{bibliography}

% -----------------------------------------------------------------------------
% REQUIRED BEFORE SaTML SUBMISSION IF THIS PAPER RECEIVED REVIEWS ELSEWHERE:
% Append the complete, anonymized, otherwise unedited reviews from the MOST
% RECENT prior submission, followed by a description of how each review was
% addressed in this SaTML version. SaTML requires this material at the very end
% of the paper, after the references and any appendices.
%
% Example structure (replace with the actual unedited anonymized reviews):
% \clearpage
% \section*{Prior Reviews and Response}
% \subsection*{Description of Changes}
% ... concise description of how the prior reviews were addressed ...
% \subsection*{Prior Review 1}
% ... complete anonymized review, otherwise unedited ...
% -----------------------------------------------------------------------------
\end{document}